\documentclass[conference]{IEEEtran}
\IEEEoverridecommandlockouts

\usepackage[utf8]{inputenc}
\usepackage[T1]{fontenc}
\usepackage{cite}
\usepackage{amsmath,amssymb,amsfonts}
\usepackage{algorithmic}
\usepackage{graphicx}
\usepackage{textcomp}
\usepackage{algorithm}
\usepackage{xcolor}
\usepackage{mathtools}
\usepackage[nolist]{acronym}
\usepackage{booktabs}
\usepackage{multirow}
\usepackage{makecell}
\usepackage{balance}
\usepackage{adjustbox} 

\begin{acronym}[XXX] 
\acro{ai}[AI]{Artificial Intelligence}
\acro{auc}[AUC]{Area Under the Curve}
\acro{cdf}[CDF]{Cumulative Distribution Function}
\acro{cnn}[CNN]{Convolutional Neural Network}
\acro{dl}[DL]{Deep Learning}
\acro{dnn}[DNN]{Deep Neural Network}
\acro{gelu}[GELU]{Gaussian Error Linear Unit}
\acro{glai}[GLAI]{GreenLightningAI}
\acro{kd}[KD]{Knowledge Distillation}
\acro{ml}[ML]{Machine Learning}
\acro{mlp}[MLP]{Multilayer Perceptron}
\acro{mva}[MVA]{Maximum Validation Accuracy}
\acro{mha}[MHA]{Multihead Attention}
\acro{moe}[MoE]{Mixture-of-Experts}
\acro{mish}[Mish]{Mish}
\acro{nlp}[NLP]{Natural Language Processing}
\acro{nn}[NN]{Neural Network}
\acro{oui}[OUI]{Overfitting-Underfitting Indicator}
\acro{pdf}[PDF]{Probability Density Function}
\acro{prelu}[PReLU]{Parametric Rectified Linear Unit}
\acro{poc}[PoC]{Proof-of-Concept}
\acro{relu}[ReLU]{Rectified Linear Unit}
\acro{silu}[SiLU]{Sigmoid Linear Unit}
\acro{sgd}[SGD]{Stochastic Gradient Descent}
\acro{svd}[SVD]{Singular Value Decomposition}
\acro{swish}[Swish]{Swish}
\acro{srelu}[SReLU]{S-shaped ReLU}
\acro{vt}[VT]{Vision Transformer}
\acro{vit}[ViT]{Vision Transformer}
\acro{wd}[WD]{Weight Decay}
\end{acronym}

\def\BibTeX{{\rm B\kern-.05em{\sc i\kern-.025em b}\kern-.08em
    T\kern-.1667em\lower.7ex\hbox{E}\kern-.125emX}}
\begin{document}

\title{$\lambda$-GELU: Learning Gating Hardness for Controlled ReLU-ization in Deep Networks}

\author{
\makebox[.33\linewidth][c]{
\begin{tabular}{c}
Cristian Pérez-Corral*\thanks{*Corresponding author} \\
\textit{Universitat Politècnica de València} \\
Valencia, Spain \\
cpercor@upv.es
\end{tabular}
}
\makebox[.33\linewidth][c]{
\begin{tabular}{c}
Alberto Fernández-Hernández \\
\textit{Universitat Politècnica de València} \\
Valencia, Spain \\
a.fernandez@upv.es
\end{tabular}
}

\makebox[.33\linewidth][c]{
\begin{tabular}{c}
Jose I. Mestre \\
\textit{Universitat Jaume I} \\
Castelló de la Plana, Spain \\
jmiravet@uji.es
\end{tabular}
}\\[1em]
\makebox[.33\linewidth][c]{
\begin{tabular}{c}
Manuel F. Dolz \\
\textit{Universitat Jaume I} \\
Castelló de la Plana, Spain \\
dolzm@uji.es
\end{tabular}
}
\makebox[.33\linewidth][c]{
\begin{tabular}{c}
Enrique S. Quintana-Ortí \\
\textit{Universitat Politècnica de València} \\
Valencia, Spain \\
quintana@disca.upv.es
\end{tabular}
}

}
\maketitle

\begin{abstract}
\ac{gelu} is a widely used smooth alternative to \ac{relu}, yet many deployment, compression, and analysis toolchains are most naturally expressed for piecewise-linear (\acs{relu}-type) networks. We study a hardness-parameterized formulation of \acs{gelu}, \(f(x;\lambda)=x\,\Phi(\lambda x)\), where \(\Phi\) is the Gaussian \ac{cdf} and \(\lambda\in[1,\infty)\) controls gate sharpness, with the goal of turning smooth gated training into a controlled path toward \acs{relu}-compatible models. Learning \(\lambda\) is non-trivial: naive updates yield unstable dynamics and effective gradient attenuation, so we introduce a constrained reparameterization and an optimizer-aware update scheme.

Empirically, across a diverse set of model--dataset pairs spanning MLPs, CNNs, and Transformers, we observe structured layerwise hardness profiles and assess their robustness under different initializations. We further study a deterministic \acs{relu}-ization strategy in which the learned gates are progressively hardened toward a principled target, enabling a post-training substitution of $\lambda$-\acs{gelu} by \acs{relu} with reduced disruption. Overall, $\lambda$-\acs{gelu} provides a minimal and interpretable knob to profile and control gating hardness, bridging smooth training with \acs{relu}-centric downstream pipelines.
\end{abstract}


\section{Introduction}
Activation functions shape both the optimization dynamics and the functional form of deep neural networks. Modern practice is largely dominated by rectified nonlinearities such as \acs{relu}~\cite{nair2010relu, pmlr-v15-glorot11a, DBLP:conf/nips/KrizhevskySH12}, and by smooth, self-gated alternatives (most notably~\acs{gelu}) that have become standard in several Transformer architectures \cite{hendrycks2016gaussian}. While smooth gates can be convenient during training, many downstream pipelines in efficient deployment and model analysis are most naturally expressed for piecewise-linear networks and often assume~\acs{relu}-type structure (e.g., model compression through pruning and quantization, and piecewise-linear interpretability and verification toolchains) \cite{jacob2018, hu2016, chu2018, katz2017}. This tension motivates mechanisms that retain the benefits of smooth training recipes while enabling a controlled transition to \acs{relu}-compatible models.

In this work, we study a hardness-parameterized variant of \acs{gelu} defined as $f(x;\lambda)=x\,\Phi(\lambda x)$, where $\Phi$ is the Gaussian \acf{cdf} and $\lambda\in[1,\infty)$ controls the sharpness of the gate. Learning $\lambda$ is non-trivial and requires a constrained parameterization together with a dedicated optimization scheme to avoid unstable updates and gradient attenuation, which we develop in this paper. Beyond enabling hardness profiling across layers, the learned parameter provides a practical control knob for \acs{relu}-ization: starting after the first $25\%$ of training epochs, we anneal $\lambda$ toward a large target and evaluate whether the resulting model can be replaced by \acs{relu} with minimal loss in the validation metric.

We emphasize that $\lambda$-\acs{gelu} plays two complementary roles in this work. First, when $\lambda$ is learned jointly with the weights, it acts as a lightweight probe of layerwise gating hardness, enabling us to characterize consistent hardness hierarchies across architectures and initializations. Second, the same parameter provides a practical control knob for \acs{relu}-ization: once an initial smooth-training phase is completed, $\lambda$ can be deterministically annealed to enforce \acs{relu}-like gating and enable a clean activation replacement for downstream compatibility or inference deployment.

In summary, our contributions are:
\begin{itemize}
    \item We introduce $\lambda$-\acs{gelu} as a minimal hardness-controlled family $x\,\Phi(\lambda x)$ with $\lambda\in[1,\infty)$.
    \item We develop a stable constrained optimization scheme to learn $\lambda$ jointly with the network parameters, based on softplus reparameterization and a dedicated learning-rate scaling for this hardness variable.
    \item We propose a deterministic hardening-and-replacement protocol: after a smooth-training phase, we freeze $\lambda$ and linearly anneal it toward a principled target $\lambda_{\mathrm{target}}$, and we evaluate \acs{relu} substitution at the best-validation checkpoint without further weight updates.
\end{itemize}

The remainder of the paper is organized as follows. Section~\ref{sec:related_work} reviews related work on smooth and parametric activation functions and on \acs{relu}-centric downstream pipelines in compression, interpretability, and verification. Section~\ref{sec:param_gelu} introduces the proposed framework and the constrained optimization strategy for learning $\lambda$. Section~\ref{sec:experiments} reports an empirical evaluation across multiple datasets and architectures and studies the proposed annealing procedure. Finally, Section~\ref{sec:conclusions} concludes and outlines future research directions.

\section{Related Work}
\label{sec:related_work}

Rectified activations, and in particular \acs{relu}, mitigate gradient attenuation effects that hinder deep optimization and have become a default choice in many pipelines~\cite{nair2010relu, pmlr-v15-glorot11a, DBLP:conf/nips/KrizhevskySH12}. At the same time, their non-differentiability at the origin and issues such as inactive (``dead'') units have motivated a large body of work on smooth alternatives. Representative examples include \acs{gelu}~\cite{hendrycks2016gaussian}, \acs{silu}~\cite{elfwing2017silu}, Swish~\cite{ramachandran2017searchingactivationfunctions}, and Mish~\cite{misra2020mishselfregularizednonmonotonic}. Many of these can be interpreted as \emph{self-gated} activations of the form $x\,\phi(x)$, where the gate $\phi(x)\in(0,1)$ is a smooth approximation to a Heaviside-type decision. In particular, \acs{gelu} has become prevalent in Transformer-based architectures, where it is frequently adopted as a default activation~\cite{hendrycks2016gaussian}.

A natural extension of both rectifiers and smooth gates is to make the nonlinearity \emph{adaptive} by introducing learnable parameters controlling slope, curvature, saturation, or sharpness. Examples include parametric rectifiers such as \acs{prelu}~\cite{he2015delving} and other variants that learn thresholds or slopes, as well as smooth gates with tunable ``temperature''-like parameters. These approaches are motivated by
improved optimization and task adaptivity, but they also raise stability concerns: unconstrained parameters may drift to degenerate regimes, and naive parameterizations can lead to vanishing or overly unstable gradients. 

The closest formulation to our own is the learnable-$\beta$ variant of Swish~\cite{ramachandran2017searchingactivationfunctions}, $x\cdot\sigma(\beta x)$, which also introduces a scalar sharpness-like parameter. However, $\beta$ is unconstrained (it can take negative values, collapsing toward zero), and no annealing protocol toward a piecewise-linear target has been proposed for it. \acs{prelu}~\cite{he2015delving} is also related in that it learns a per-layer scalar, but it modifies the slope of the negative half-plane rather than gating sharpness, and does not converge to \acs{relu} as the parameter grows. Our $\lambda$-\acs{gelu} differs from both in three concrete ways: (i)~$\lambda\ge 1$ is enforced by construction, grounding the family at standard \acs{gelu}; (ii)~we derive a principled annealing target $\lambda_{\mathrm{target}}$ from a closed-form $\ell_1$ approximation bound between $\Phi(\lambda x)$ and the Heaviside gate; and (iii)~we directly evaluate \acs{relu} substitution quality at the end of training. Our work, thus, contributes to this line by explicitly modeling a hardness parameter in a \acs{gelu}-style gate and by focusing on how to learn it stably under the natural constraint $\lambda\ge 1$ via reparameterization and optimizer-aware scaling. 

Beyond training-time considerations, a large ecosystem of deployment and analysis methods is best aligned with piecewise-linear networks and often implicitly assumes \acs{relu}-type nonlinearities. This is particularly clear in:

\noindent\textbf{Quantization and efficient inference.}
Integer-only quantization pipelines are most naturally expressed with linear operators and simple piecewise-linear nonlinearities, and classic schemes for quantization-aware training and efficient inference are commonly presented for \acs{relu}-based networks~\cite{jacob2018}. In contrast, smooth nonlinearities such as \acs{gelu} require either mixed-precision handling or dedicated approximations to enable efficient integer-only execution. This challenge is explicit in work on quantized Transformers, where \acs{gelu} is approximated to support integer arithmetic~\cite{kim2021bert}.

\noindent\textbf{Pruning and sparsity-driven compression.}
Many pruning and compression heuristics exploit the exact zeroing behavior induced by rectifiers, enabling criteria based on activation sparsity and facilitating structured removal of units or channels. For instance, Network Trimming leverages the Average Percentage of Zeros (APoZ) after \acs{relu} to identify and prune redundant neurons/feature maps~\cite{hu2016}. Such sparsity-based signals are less direct for smooth activations that rarely produce exact zeros, which can reduce both the interpretability of sparsity metrics and potential hardware speedups.

\noindent\textbf{Interpretability and formal analysis for piecewise-linear networks.}
Several ``white-box'' interpretability methods are designed around piecewise-linear structure, enabling exact region-wise linear explanations of network behavior. \textsc{OpenBox}, for example, provides exact and consistent interpretability for piecewise-linear neural networks by converting them into equivalent collections of linear models~\cite{chu2018}. Similarly, formal verification and robustness analysis toolchains commonly target \acs{relu} networks because \acs{relu} constraints admit efficient reasoning via SAT/SMT-style procedures, linear relaxations, or dedicated solvers. Reluplex~\cite{katz2017} and Marabou~\cite{katz2019} are representative systems that build on \acs{relu}'s piecewise-linear semantics to verify properties of neural networks.

Motivated by these strands, we study a hardness-parameterized \acs{gelu} gate that can be learned during training and subsequently annealed to approach \acs{relu}-like behavior. This provides a practical bridge between smooth-gated training (often preferred in modern architectures) and \acs{relu}-compatible downstream pipelines in quantization, pruning, interpretability, and verification.

\section{Parametrized \acs{gelu}}
\label{sec:param_gelu}
Let $\Phi$ and $\varphi$ denote the \ac{cdf} and \ac{pdf} of the standard normal distribution $\mathcal{N}(0,1)$. We study a parametric variant of \acs{gelu} defined as
\begin{equation}
\label{eq:lgelu_scalar}
\lambda\text{-}\mathrm{GELU}(x) \;=\; x\,\Phi(\lambda x),
\end{equation}
where $\lambda \ge 1$ controls the sharpness of the gate. Note that the standard \acs{gelu} is recovered at $\lambda=1$, since $\mathrm{GELU}(x)=x\Phi(x)$.

The gating term $\Phi(\lambda x)$ becomes increasingly sharp as $\lambda$ grows. In particular, as $\lambda\to\infty$ we have $\Phi(\lambda x)\to 1$ when $x>0$ and $\Phi(\lambda x)\to 0$ when $x<0$. Therefore,
\begin{equation}
\label{eq:relu_limit}
\lim_{\lambda\to\infty} x\,\Phi(\lambda x)
\;=\;
x\,H(x)
\;=\;
\mathrm{ReLU}(x)
\end{equation}
where $H$ denotes the Heaviside step function, with $H(0) = \frac{1}{2}$. Consequently, $\lambda$-\acs{gelu} interpolates between \acs{gelu}-like smooth gating (small $\lambda$) and \acs{relu}-like hard gating (large $\lambda$). This parameter $\lambda$ can be shared globally, per layer, per channel, or per neuron, although in this work we only consider the layerwise scenario (one shared $\lambda_\ell$ per activation layer).

Optimizing $\lambda\in[1,\infty)$ directly can be inconvenient for SGD-like methods because unconstrained gradient steps may propose updates that cross below $1$. We thus introduce an unconstrained parameter $s\in\mathbb{R}$ and map it to $\lambda>1$ through a softplus reparameterization:
\[
\label{eq:reparam}
\lambda(s) \;=\; 1 + \mathrm{softplus}\!\left(\frac{s}{t}\right) = 1 + \log(1+e^{\frac{s}{t}}),
\]
where $t>0$ is a temperature controlling the sensitivity of $\lambda$ with respect to $s$. In this work, $t$ is treated as a hyperparameter (not learned). 
This reparameterization has three practical benefits. First, it converts the constrained optimization problem $\lambda\in[1,\infty)$ into unconstrained updates over $s\in\mathbb{R}$, which is directly compatible with standard SGD-like optimizers. Second, the mapping is smooth and computationally inexpensive, as softplus is optimized in most \ac{dl} frameworks. Third, it makes explicit how the temperature $t$ modulates the sensitivity of $\lambda$ to gradient-based updates.

Let $\delta := \partial L/\partial a$ denote the backpropagated signal at the activation output $a(x;\lambda)=x\Phi(\lambda x)$. By the chain rule,
\[
\frac{\partial L}{\partial s}
=
\frac{\partial L}{\partial a}\,
\frac{\partial a}{\partial \lambda}\,
\frac{\partial \lambda}{\partial s}
=
\delta \cdot x^2\,\varphi(\lambda x)\cdot
\frac{1}{t}\,\sigma\!\left(\frac{s}{t}\right),
\]
where $\varphi$ is the standard normal \ac{pdf} and $\sigma$ is the sigmoid function. Having
\(
s_{k+1} \;=\; s_k - \eta_s \,\frac{\partial L}{\partial s},
\)
as the change in $s$, the induced change in $\lambda$ can be approximated to first order as
\begin{equation}
\begin{split}
\label{eq:effective_lambda_step}
\Delta \lambda
&=
\lambda(s_{k+1})-\lambda(s_k)
\approx
\frac{\partial \lambda}{\partial s}\,\Delta s \\
&=
-\eta_s\left(\frac{\partial \lambda}{\partial s}\right)^2
\frac{\partial L}{\partial \lambda}
=
-\eta_s\frac{\sigma(s/t)^2}{t^2}\,\frac{\partial L}{\partial \lambda}.
\end{split}
\end{equation}
Equation~\eqref{eq:effective_lambda_step} highlights two effects. The effective step size on $\lambda$ scales as $1/t^2$, but it is also \emph{state-dependent} through $\sigma(s/t)$: the same learning rate $\eta_s$ can translate into very different $\Delta\lambda$ depending on the current value of $s$. Therefore, changing $t$ is not equivalent to a mere constant rescaling of the learning rate; it also alters the local sensitivity of the mapping $\lambda(s)$ and the magnitude of gradients flowing through $s$.

To control the adaptation speed of the hardness parameter relative to the network weights, we introduce a learning-rate multiplier. Let $\eta_w$ denote the learning rate used for the weights, and set $\eta_s =c\cdot\eta_w$, where $c>0$ is a hyperparameter controlling how aggressively $\lambda$ adapts compared to the weights. In practice, the effect of this scaling depends on the optimizer: under SGD, $c$ acts as a direct rescaling of the update magnitude, whereas under Adam-like adaptive methods the update is additionally shaped by first- and second-moment estimates, so overly large $c$ may lead to unstable hardness dynamics. In Section~\ref{sec:experiments}, we tune $(t,c)$ on a small grid and report stable, best-performing configurations, which we then reuse across the considered model--dataset pairs.

In this work we learn $\lambda$ layerwise (one hardness parameter per layer), and interpret $\lambda_\ell$ as a probe of how decisive the gating should be in layer $\ell$: smaller values correspond to smoother transitions, while larger values yield increasingly \acs{relu}-like behavior. Empirically, we first characterize the resulting layerwise hardness profiles across architectures and study their robustness across random initializations.

Motivated by the fact that many downstream pipelines are most naturally expressed for \acs{relu}/piecewise-linear networks, we propose a gradual hardening procedure. Specifically, we train $\lambda$-\acs{gelu} normally for the first $25\%$ of the training epochs and then progressively increase $\lambda$ according to a prescribed annealing schedule over the remaining epochs. We detail the protocol and report results in Section~\ref{sec:experiments}.

\section{Experiments}
\label{sec:experiments}

This section reports the empirical results of our approach and evaluates it under different experimental conditions.

As stated above, this work does not aim to introduce a new activation function. Rather, we use the $\lambda$-\acs{gelu} parametrization as a probe to study how gating hardness evolves during training, by tracking the learned sharpness parameter across depth. Accordingly, our experimental goal is not to outperform the baselines in predictive performance, but (i) to characterize and compare the resulting hardness dynamics across architectures and datasets; and (ii) to evaluate a hardening-and-replacement procedure that anneals $\lambda$ after the first $25\%$ of training epochs and tests whether the resulting model can be replaced by a \acs{relu} network with minimal loss in validation performance. 

To this end, we selected a diverse set of models and datasets to cover the main architectural families used in contemporary \ac{dl}. For each family, we previously perform a hyperparameter sweep over the learning rate (LR) and weight decay (WD) on the baseline model, and report only the configuration achieving the highest validation score. We consider:
\begin{itemize}
    \item \textbf{\acp{mlp}:} a 4-layer \ac{mlp} with 256 hidden units per layer, trained with SGD using $\mathrm{LR}=5\cdot 10^{-2}$ and $\mathrm{WD}=10^{-4}$.
    \item \textbf{\acp{cnn}:} ResNet-18, trained with AdamW using $\mathrm{LR}=5\cdot 10^{-4}$ and $\mathrm{WD}=10^{-4}$.
    \item \textbf{Transformers:} DeiT (vision) and GPT-2 (language). DeiT is trained with AdamW using $\mathrm{LR}=5\cdot 10^{-4}$ and $\mathrm{WD}=7\cdot 10^{-2}$, while GPT-2 is trained with AdamW using $\mathrm{LR}=5\cdot 10^{-5}$ and $\mathrm{WD}=10^{-2}$.
\end{itemize}
We evaluate six model--dataset benchmarks: \acs{mlp}/Adult~\cite{adult}, \acs{mlp}/FMNIST~\cite{fmnist}, ResNet-18~\cite{resnet}/CIFAR-10\cite{cifar10}, ResNet-18~\cite{resnet}/CIFAR-100~\cite{cifar10}, DeiT (Tiny)~\cite{deit} /TinyImageNet~\cite{tiny}, and GPT-2~\cite{gpt2}/WikiText-2\cite{wikitext}. All experiments are conducted in Python~v3.10 with PyTorch~v2.6; we fix $3$ random seeds for reproducibility.
Metrics are reported as mean $\pm$ standard deviation across the $3$ seeds. We use validation accuracy for tabular/vision benchmarks and validation perplexity (lower is better) for GPT-2 on WikiText-2.

Unless stated otherwise, we report averages over $3$ fixed random seeds. ``Layerwise'' hardness means one shared parameter $\lambda_\ell$ per layer (one scalar per activation layer). This same $\lambda_\ell$ is used by the activation applied to all units in \ac{mlp} layers, all channels in \ac{cnn} activation sites, and all neurons in Transformer FFN sublayers. For optimization, the unconstrained parameters $s$ use a separate learning rate $\eta_s=c\,\eta_w$. These parameters are also excluded from weight decay (only the weights are regularized). Additional implementation details and full configurations are provided in the repository\footnote{All code configurations will be provided upon acceptance}.

\subsection{Tuning the reparameterization hyperparameters $(t,c)$}
The first experiment performs a grid search over the softplus temperature $t$ and the multiplier $c$ that scales the learning rate of the unconstrained parameter $s$ (cf. Section~\ref{sec:param_gelu},  $\eta_s=c\,\eta_w$). Concretely, we evaluate
\[
(t,c)\in\{0.1,\,0.3,\,0.6,\,0.9\}\times\{1,\,3,\,6,\,9\},
\]
using an \ac{mlp} trained on FMNIST with SGD. We repeat each configuration under three initializations of $s$ and three initialization modes $\mathcal{M}=\{\text{uniform},\text{increasing},\text{decreasing}\}$, which control how $s$ (and thus $\lambda$) is initialized across depth. The uniform mode sets the same $s$ in every layer, yielding $\lambda \approx 1$ throughout. The increasing/decreasing modes initialize $\lambda$ monotonically across layers within the range $[1,2]$ (from $\approx 1$ to $\approx 2$, or vice versa) to test whether the learned hardness allocation is robust to the initial ordering. We restrict this range to stay close to the $\lambda{=}1$ (\ac{gelu}) baseline while still inducing distinct depth orderings: larger initial values would partially harden the network from the start and substantially attenuate the learning signal for $\lambda$, since $\partial a/\partial \lambda = x^2\varphi(\lambda x)$ concentrates around $x=0$ as $\lambda$ grows, leading to small and potentially high-variance gradients under typical pre-activation scales. For a run identified by random seed $r$, initialization mode $m\in\mathcal{M}$, temperature $t$, and multiplier $c$, let $\lambda_{\ell}^{(r,m,t,c)}(e)$ denote the learned hardness in layer $\ell\in\{1,\dots,L\}$ at epoch $e$.
We quantify hardness adaptation via the average per-layer drift between consecutive epochs,
\[
V_{\lambda}(r,m,t,c)=\frac{1}{L}\sum_{\ell=1}^{L}\sum_{e=1}^{T-1}
\Big|\lambda_{\ell}^{(r,m,t,c)}(e+1)-\lambda_{\ell}^{(r,m,t,c)}(e)\Big|,
\]
and report its cell-average
\[
\Delta\lambda(t,c)=\mathbb{E}_{r}\,\mathbb{E}_{m\in\mathcal{M}}\big[\,V_{\lambda}(r,m,t,c)\,\big].
\]
$\Delta\lambda(t,c)$ should be interpreted as a proxy for ``how much hardness adapts'' under a given $(t,c)$, rather than as an absolute measure of final gate sharpness or a direct indicator of predictive performance. To assess the latter, we use the best validation score (BVS) reached during training. Let
\[
\mathrm{BVS}(r,m,t,c)=\max_{e\in\{1,\dots,T\}}\mathrm{val}^{(r,m,t,c)}(e)
\]
denote the best validation score of a run. The mean best validation score for $\lambda$-\acs{gelu} and the \acs{gelu} baseline are
\begin{equation}
\begin{split}
\overline{\mathrm{BVS}}_{\lambda}(t,c)
&=\mathbb{E}_{r}\,\mathbb{E}_{m\in\mathcal{M}}\!\left[\mathrm{BVS}(r,m,t,c)\right],
\\
\overline{\mathrm{BVS}}_{\mathrm{GELU}}
&=\mathbb{E}_{r}\!\left[\mathrm{BVS}(r,\mathrm{GELU})\right],
\end{split}
\end{equation}
and the heatmap annotations report their difference,
\[
\Delta\mathrm{BVS}(t,c)=\overline{\mathrm{BVS}}_{\lambda}(t,c)-\overline{\mathrm{BVS}}_{\mathrm{GELU}}.
\]

\begin{figure}[htpb]
    \centering
    \includegraphics[width=\linewidth]{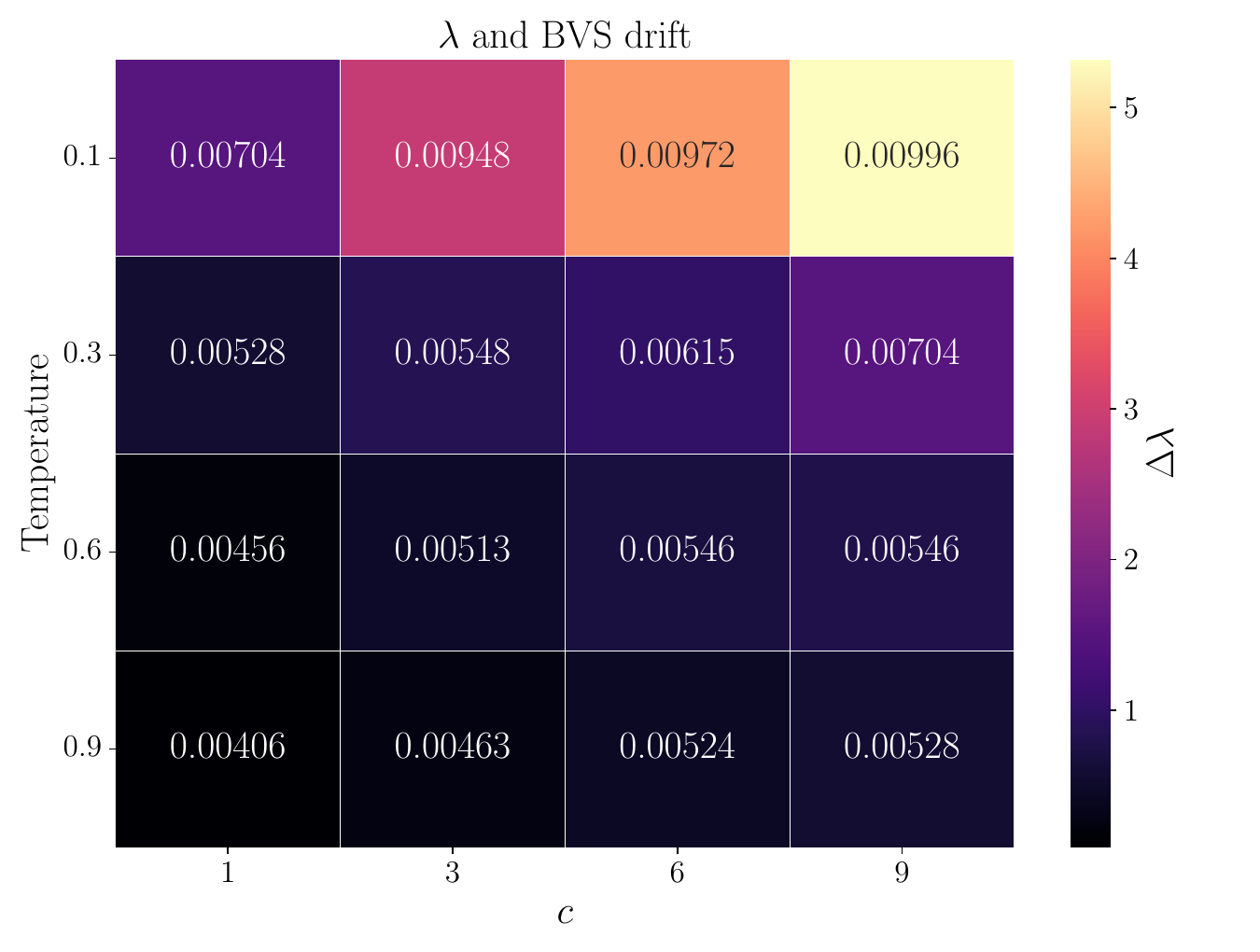}\vspace{-0.3cm}
    \caption{Grid over temperature $t$ and the $s$ learning-rate multiplier $c$ on \ac{mlp}/FMNIST. Cell annotations report the mean validation-score change with respect to \acs{gelu} (averaged over the three $s$ initializations), while the color encodes the mean hardness-drift proxy $\Delta\lambda$ (average absolute epoch-to-epoch change of the learned layerwise hardness). Low temperatures and larger $c$ induce stronger hardness adaptation.}
    \label{fig:heatmap}
\end{figure}

Figure~\ref{fig:heatmap} summarizes the grid: the color encodes the mean hardness drift $\Delta\lambda(t,c)$, while the cell annotation shows $\Delta\mathrm{BVS}(t,c)$. Two expected monotonic trends are visible. First, smaller temperatures increase the sensitivity of the mapping $\lambda(s)$, amplifying hardness adaptation. Second, larger $c$ yields larger updates on $s$, which also increases $\Delta\lambda$. Among the tested temperatures, $t=0.1$ provides the most favorable trade-off, retaining substantial hardness drift while keeping $\Delta\mathrm{BVS}$ close to the \acs{gelu} baseline.

\begin{figure}[hptb]
    \centering
    \includegraphics[width=\linewidth]{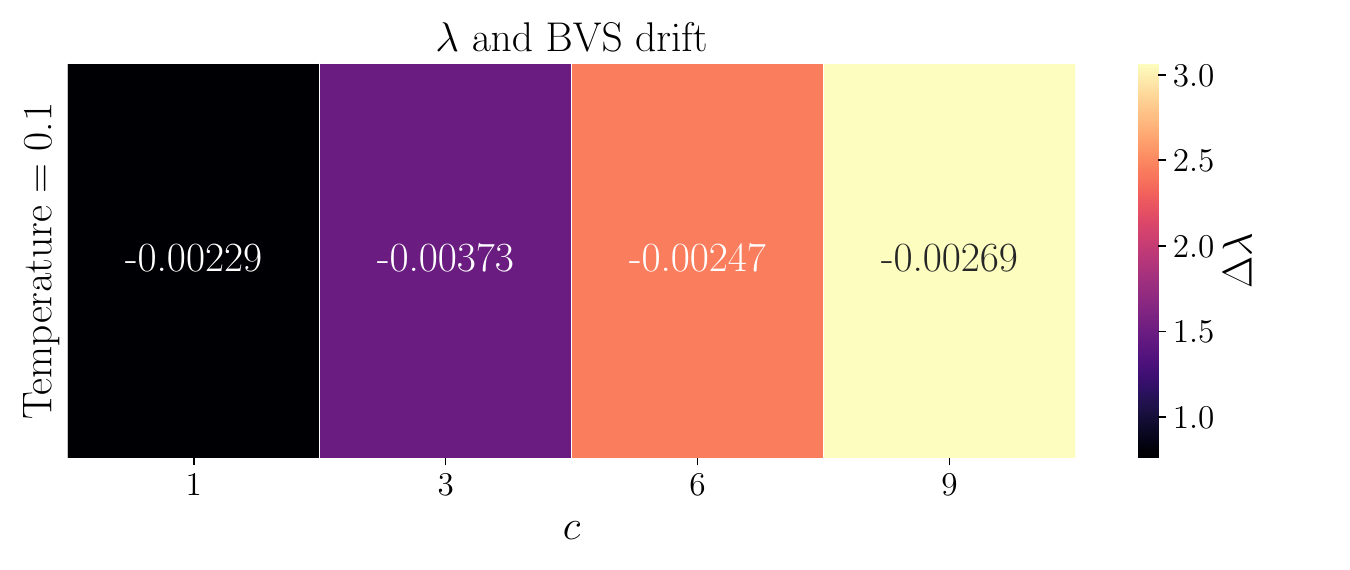}\vspace{-0.3cm}
    \caption{Sweep over the $s$ learning-rate multiplier $c$ on ResNet-18/CIFAR-100 (AdamW, $t{=}0.1$ fixed). Cell annotations report the mean validation-score change with respect to \acs{gelu} (averaged over the three $s$ initializations), while the color encodes the mean hardness-drift proxy $\Delta\lambda$ (average absolute epoch-to-epoch change of the learned layerwise hardness). Larger $c$ monotonically increases $\Delta\lambda$ while $\Delta\mathrm{BVS}$ remains small across the sweep; $c{=}1$ achieves the best mean $\Delta\mathrm{BVS}$ but $c{=}9$ is comparable while maximizing hardness adaptation.}
    \label{fig:heatmap2}
\end{figure}

Because scaling $\eta_s$ under AdamW is not equivalent to a pure gradient rescaling (due to first/second-moment accumulation), we additionally run a reduced proxy (ResNet18 on CIFAR100) sweep for AdamW-based settings: we fix $t=0.1$ and sweep $c\in\{1,3,6,9\}$. As shown in Figure~\ref{fig:heatmap2}, increasing $c$ again monotonically increases $\Delta\lambda$, while $\Delta\mathrm{BVS}$ remains small in magnitude across the sweep. Although $c=1$ achieves the best mean $\Delta\mathrm{BVS}$ in this proxy, $c=9$ is comparable while maximizing hardness adaptation.

Overall, across the explored grid, learning $\lambda$ leaves the best validation score close to the \acs{gelu} baseline while enabling a controllable amount of hardness adaptation. In our subsequent experiments we fix $t=0.1$ and $c=9$ as a simple choice that consistently produced substantial $\Delta\lambda$ without degrading BVS in the considered settings; we do not claim this pair to be universally optimal across all architectures and optimizers.


\subsection{Evaluation of $\lambda$-\acs{gelu} under different initializations}
\begin{figure}[htpb]
    \centering
    \includegraphics[width=0.93\linewidth]{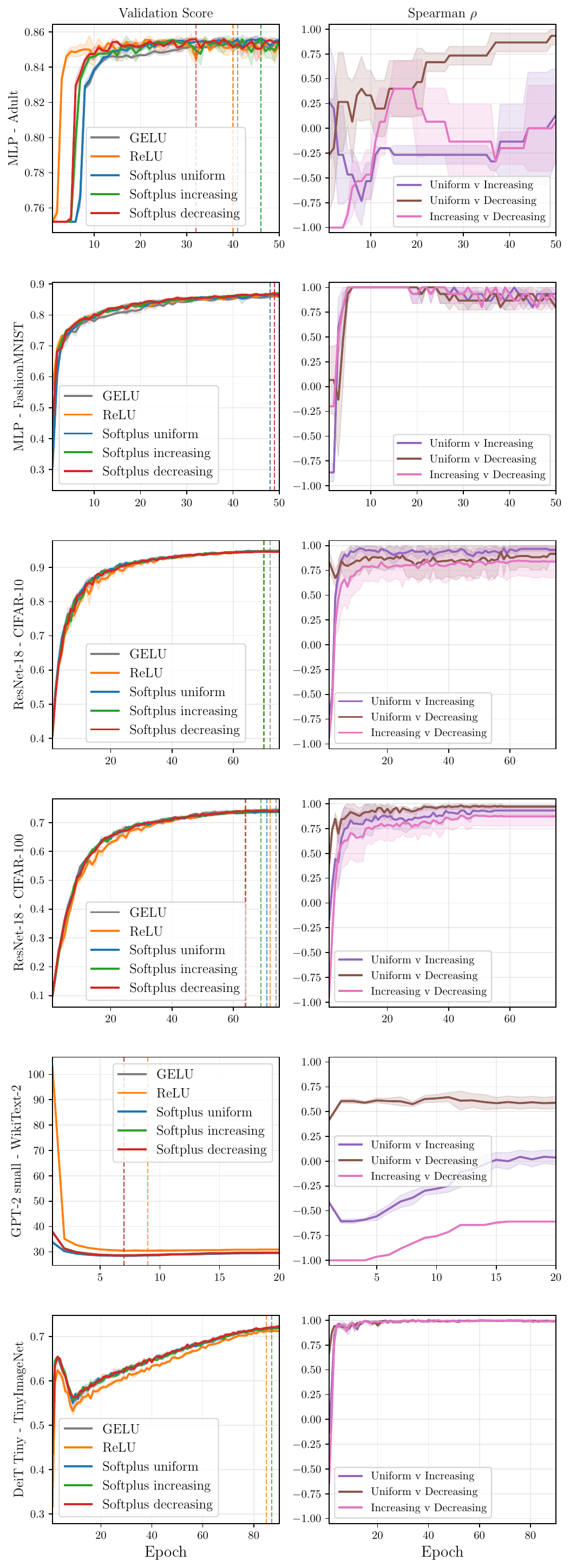}\vspace{-0.3cm}
    \caption{Validation-metric trajectories (left) and Spearman rank correlations between layerwise hardness profiles across $s$-initialization modes (right). Vertical lines mark the epoch of best validation score (BVS).}
    \label{fig:spearman_validation}
\end{figure}

The next experiment evaluates all selected model--dataset pairs and characterizes how the learned layerwise hardness parameters evolve throughout training. For each configuration, we report: (i) the validation-metric trajectory; and (ii) the Spearman rank correlation between the layerwise hardness vectors obtained under different initializations of the parameter $s$. At each epoch $e$, we form the layerwise hardness profile $\bar{\boldsymbol{\lambda}}^{(r,m)}(e)\in\mathbb{R}^L$ and compute the Spearman rank correlation between profiles obtained from two $s$-initialization modes $m_1,m_2$:
\[
\rho_S(e;m_1,m_2)=\mathbb{E}_r\Big[\mathrm{Spearman}\big(\bar{\boldsymbol{\lambda}}^{(r,m_1)}(e),\bar{\boldsymbol{\lambda}}^{(r,m_2)}(e)\big)\Big].
\]

High values of $\rho_S$ indicate that the relative ordering of layers by hardness is preserved across initializations, i.e., layers that become harder under one initialization also tend to be harder under the others (up to a monotone transformation). Therefore, this experiment tests whether training induces an initialization-robust layerwise hardness hierarchy, rather than an initialization-dependent allocation of hardness across depth.

Figure~\ref{fig:spearman_validation} reports validation trajectories (left) and the corresponding Spearman correlations of layerwise hardness rankings across initialization modes (right). Shadow indicates the standard deviation across seeds. Several vision settings (\ac{mlp} on \textsc{FMNIST}, ResNet-18 on CIFAR-10/100, and DeiT on TinyImageNet) converge to consistently high \(\rho_S\), indicating that training induces an initialization-robust ordering of layers by hardness. 

In contrast, for Adult and especially for GPT-2/WikiText-2, \(\rho_S\) is lower and/or stabilizes later, suggesting that under our fixed \((t,c)\) the allocation of hardness across depth is more sensitive to initialization and optimizer dynamics. This points to the need for architecture-specific tuning of hardness learning (or longer training) to obtain robust layerwise profiles in these regimes.

These results are also summarized in Table~\ref{tab:exp3} Here, BVS denotes the best validation score achieved during training: validation accuracy for all vision/tabular benchmarks, and validation perplexity for GPT-2 on WikiText-2 (lower is better). Ep. is the epoch at which BVS is reached. $\overline{\lambda}$ is the network-wide mean of the learned layerwise hardness values, and $\overline{\Delta\lambda}$ is the corresponding mean drift across seeds and layers. A mild trend can be observed: architectures that are more commonly trained with \acs{relu} (e.g., the \acs{mlp} and \acs{cnn} considered here) tend to learn larger average values of $\lambda$ than architectures typically trained with \acs{gelu}. We interpret this as a possible indication of a preference toward harder gating in those settings under our training configuration, rather than as a universal property.

Overall, Table~\ref{tab:exp3} shows that the baseline and the different $\lambda$-\acs{gelu} initialization modes achieve similar best validation scores in most benchmarks, with differences that are generally small. In our experiments, the uniform initialization is a reasonable default choice: it performs comparably to the increasing/decreasing schemes while remaining initialization-neutral and adding only one learnable hardness parameter per activation layer, which is negligible relative to the total number of network parameters.

\begin{table}[htpb]
\caption{Summary of the metrics reported in our experiments (identified by dataset).}
\label{tab:exp3}
\setlength{\tabcolsep}{5pt}

\begin{adjustbox}{max width=\linewidth}
\begin{tabular}{llccccc}
\toprule
\textbf{Dataset} & \textbf{Act.} & \textbf{Mode} & \textbf{BVS} & \textbf{Ep.} &
\makecell{\textbf{$\overline{\lambda}$}} &
\makecell{\textbf{$\overline{\Delta \lambda}$}} \\
\midrule

\multirow{7}{*}{\textbf{Adult}}
& \acs{gelu}                              & -   & $0.85  \pm4\cdot10^{-4}$         & 41           & -  & -   \\
\cmidrule(lr){2-7}
& \acs{relu}                              & -   & $0.86 \pm 6\cdot10^{-4}$ & 40           & -  & -   \\
\cmidrule(lr){2-7}
& \multirow{4}{*}{$\lambda$-\acs{gelu}} & dec & $\mathbf{0.87 \pm8\cdot10^{-4}}$& \textbf{32}  & $6.66 \pm 0.13$&   $8.10\pm$ 0.2\\
\cmidrule(lr){3-7}
&                                       & inc & $0.86 \pm 9\cdot10^{-4}$& 46           & $8.67 \pm 0.10$& $11.63 \pm 0.49$\\
\cmidrule(lr){3-7}
&                                       & uni & $0.86 \pm 10^{-3} $& 46           & $2.45 \pm 0.07$& $3.23 \pm 0.49$\\
\midrule

\multirow{7}{*}{\textbf{FMNIST}}
& \acs{gelu}                              & -   & $0.86 \pm 2\cdot 10^{-3}$& 49           & -  & -   \\
\cmidrule(lr){2-7}
& \acs{relu}                              & -   & $0.86 \pm 1\cdot 10^{-3}$& \textbf{48}  & -  & -   \\
\cmidrule(lr){2-7}
& \multirow{4}{*}{$\lambda$-\acs{gelu}} & dec & $\mathbf{0.87 \pm 3\cdot10^{-3}}$& 49           & $3.40 \pm 3\cdot10^{-2}$& $6.34\pm 0.17$\\
\cmidrule(lr){3-7}
&                                       & inc & $\mathbf{0.87 \pm 2\cdot10^{-3}}$& 50           & $3.41 \pm 3\cdot10^{-2}$& $6.85\pm 0.14$\\
\cmidrule(lr){3-7}
&                                       & uni & $\mathbf{0.87 \pm 8\cdot10^{-4}}$& \textbf{48}  & $3.09 \pm 2\cdot10^{-2}$& $6.48 \pm 0.17$\\
\midrule

\multirow{7}{*}{\textbf{CIFAR10}}
& \acs{gelu}                              & -   & $\mathbf{0.95 \pm 1 \cdot10^{-3}}$& 72           & -  & -   \\
\cmidrule(lr){2-7}
& \acs{relu}                              & -   & $\mathbf{0.95 \pm 1 \cdot10^{-3}}$& \textbf{70}  & -  & -   \\
\cmidrule(lr){2-7}
& \multirow{4}{*}{$\lambda$-\acs{gelu}} & dec & $\mathbf{0.95 \pm 2 \cdot10^{-3}}$& 75           & $2.43\pm 6\cdot10^{-2}$& $4.55\pm 0.15$\\
\cmidrule(lr){3-7}
&                                       & inc & $\mathbf{0.95 \pm 4 \cdot10^{-4}}$& \textbf{70}  & $1.68\pm 0.39$& $3.51\pm 0.76$\\
\cmidrule(lr){3-7}
&                                       & uni & $\mathbf{0.95 \pm 6 \cdot10^{-4}}$& 75           & $1.75 \pm 0.51$& $3.22\pm 0.61$\\
\midrule

\multirow{7}{*}{\textbf{CIFAR100}}
& \acs{gelu}                              & -   & $\mathbf{0.74 \pm 5\cdot 10^{-3}}$& 74           & -  & -   \\
\cmidrule(lr){2-7}
& \acs{relu}                              & -   & $0.73 \pm 3\cdot 10^{-3}$& 72           & -  & -   \\
\cmidrule(lr){2-7}
& \multirow{4}{*}{$\lambda$-\acs{gelu}} & dec & $\mathbf{0.74 \pm 3\cdot 10^{-3}}$& \textbf{64}  & $2.64\pm9\cdot10^{-2}$& $4.23 \pm 0.13$\\
\cmidrule(lr){3-7}
&                                       & inc & $\mathbf{0.74 \pm 3\cdot 10^{-3}}$& 69           & $1.65 \pm 0.48$& $3.18 \pm 0.77$\\
\cmidrule(lr){3-7}
&                                       & uni & $\mathbf{0.74 \pm 3\cdot 10^{-3}}$& 71           & $1.89 \pm 0.39$& $3.39 \pm  0.60$\\
\midrule

\multirow{7}{*}{\textbf{TinyImageNet}}
& \acs{gelu}                              & -   & $\mathbf{0.72 \pm 1\cdot 10^{-3}}$& 90           & -  & -   \\
\cmidrule(lr){2-7}
& \acs{relu}                              & -   & $0.71 \pm 8\cdot 10^{-4}$& \textbf{85}  & -  & -   \\
\cmidrule(lr){2-7}
& \multirow{4}{*}{$\lambda$-\acs{gelu}} & dec & $\mathbf{0.72 \pm 2\cdot 10^{-3}}$& 90           & $1.21 \pm 1\cdot 10^{-3}$& $1.39 \pm 2\cdot 10^{-2}$\\
\cmidrule(lr){3-7}
&                                       & inc & $\mathbf{0.72 \pm 2\cdot 10^{-3}}$& \textbf{87}  & $1.21 \pm 1\cdot 10^{-3}$& $1.43 \pm 4\cdot 10^{-2}$\\
\cmidrule(lr){3-7}
&                                       & uni & $\mathbf{0.72 \pm 1\cdot 10^{-4}}$& 90           & $1.19 \pm 2\cdot 10^{-4}$& $1.36 \pm 3\cdot 10^{-2}$\\
\midrule

\multirow{7}{*}{\textbf{WikiText-2}}
& \acs{gelu}                              & -   & $28.46\pm 2\cdot 10^{-2}$& 7            & -  & -   \\
\cmidrule(lr){2-7}
& \acs{relu}                              & -   & $30.48\pm 4\cdot 10^{-2}$& 9            & -  & -   \\
\cmidrule(lr){2-7}
& \multirow{4}{*}{$\lambda$-\acs{gelu}} & dec & $28.59\pm 9\cdot 10^{-3}$& 7            & $1.24 \pm 6 \cdot 10^{-5}$& $0.28 \pm 2 \cdot 10^{-4}$\\
\cmidrule(lr){3-7}
&                                       & inc & $28.71\pm 2\cdot 10^{-2}$& 7            & $1.24 \pm 1 \cdot 10^{-4}$& $0.29 \pm 2 \cdot 10^{-5}$\\
\cmidrule(lr){3-7}
&                                       & uni & $\mathbf{28.45\pm 2\cdot 10^{-2}}$& 7            & $1.00 \pm 8 \cdot 10^{-6}$& $2\cdot 10^{-3} \pm 1 \cdot 10^{-5}$\\
\bottomrule
\end{tabular}
\end{adjustbox}\end{table}

\subsection{Hardness annealing and \acs{relu} replacement}  
A practical motivation for learning gating hardness is to enable a \emph{controlled} conversion from \acs{gelu}-based training to a \acs{relu}-compatible model at the end of training. Rather than retraining a \acs{relu} network from scratch or performing a hard swap mid-training (which may introduce an abrupt distributional shift), we aim to progressively sharpen the learned gates. This way, the resulting network approaches piecewise-linear behavior and becomes amenable to downstream pipelines that assume \acs{relu}/piecewise-linear structure (e.g., quantization, sparsity-driven compression, and analysis toolchains).

In this experiment, we propose a gradual hardening procedure followed by a post-training activation substitution test. We train $\lambda$-\acs{gelu} normally for the first $25\%$ of epochs, and then enforce a deterministic hardening schedule. We set the switch point to $25\%$ of training as a reasonable fixed baseline that preserves an initial smooth-gating phase while leaving sufficient training time for the network to adapt to the progressive hardening. Switching too early (e.g., $10\%$) shortens the smooth-gating phase and may underuse the optimization benefits of \acs{gelu}-like curvature, whereas switching too late (e.g., $50\%$) may increase the abruptness of the distribution shift induced by hardening. Using a fixed $25\%$ switch across all benchmarks also avoids per-dataset tuning and provides a consistent evaluation protocol. To evaluate replacement, we select the checkpoint corresponding to the best validation epoch (under the original activation or annealed one), and at that checkpoint we measure: (i) the validation metric with the original activation; and (ii) the validation metric after substituting \acs{gelu}/$\lambda$-\acs{gelu} by \acs{relu}, without any further weight updates and on the same validation split.

\begin{algorithm}[htpb]
\caption{Deterministic hardening and ReLU substitution}
\label{alg:anneal_swap}
\begin{algorithmic}[1]
\STATE Train $\lambda$-\acs{gelu} with learnable $\{s_\ell\}_{\ell=1}^L$ for $e=1,\dots,e_s=\lfloor 0.25T\rfloor$.
\STATE Save the learned hardness values $\{\lambda_\ell(e_s)\}_{\ell=1}^L$.
\FOR{$e=e_{s+1},\dots,T$}
\STATE Set $\lambda_\ell(e)\leftarrow \lambda_\ell(e_s) + \frac{e-e_s}{T-e_s}\left(\lambda_{\mathrm{target}}-\lambda_\ell(e_s)\right)$ for all $\ell$.
\STATE Freeze $s$ (no gradient updates) and continue training weights.
\ENDFOR
\STATE Select the best-validation checkpoint (under $\lambda$-\acs{gelu}).
\STATE Evaluate validation metric with $\lambda$-\acs{gelu}; then substitute activations by ReLU and re-evaluate.
\end{algorithmic}
\end{algorithm}
In all annealing runs, the schedule enforces $\lambda_\ell(T)=\lambda_{\mathrm{target}}$ for every layer $\ell$, ensuring that the final gates are uniformly hardened before the substitution test. To clarify the annealing process, we present it as pseudocode in Algorithm~\ref{alg:anneal_swap}. Note that during hardening, $\lambda$ is not learnable: we stop gradient updates for $s$ and override $\lambda$ with a deterministic linear schedule toward $\lambda_{\mathrm{target}}$ over the remaining $75\%$ of training. Algorithm~\ref{alg:anneal_swap} summarizes the procedure. To set the annealing target $\lambda_{\mathrm{target}}$ in a principled way, we quantify how well the smooth gate $\Phi(\lambda x)$ approximates the hard gate $H(x)$ as a function of $\lambda$. A closed-form computation yields the global bound
\[
E_{\infty}(\lambda)=\int_{-\infty}^{\infty}\!\left|H(x)-\Phi(\lambda x)\right|dx=\frac{2}{\lambda\sqrt{2\pi}}.
\]
Therefore, given a tolerance $\varepsilon>0$, it suffices to choose $\lambda_{\mathrm{target}}\ge \frac{2}{\varepsilon\sqrt{2\pi}}$;  in particular, $\varepsilon=5\times 10^{-3}$ gives $\lambda_{\mathrm{target}}\approx160$.
In the experiments below, we use this criterion to select $\lambda_{\mathrm{target}}$ for annealing, and we report results across all model--dataset benchmarks by comparing the \acs{gelu} baseline and the $\lambda$-\acs{gelu} with post-$25\%$ hardness annealing toward $\lambda_{\mathrm{target}}$.

We use $E_\infty(\lambda)$ to choose $\lambda_{\mathrm{target}}$ because it is distribution-agnostic: it depends only on the gate family and the desired tolerance, and it does not require estimating pre-activation statistics that vary across architectures due to normalization (e.g., batch/layer normalization) and training dynamics. This makes $\lambda_{\mathrm{target}}$ a conservative but generic choice that can be reused across all model--dataset pairs. 

We report annealing-and-substitution results for the tabular and vision benchmarks. For GPT-2/WikiText-2, activation substitution remains unstable under our protocol (see discussion below), so we omit it from Table~\ref{tab:eval}.
Table~\ref{tab:eval} reports these experiments, where the $\lambda$-\acs{gelu} has been annealed. In particular, we select the checkpoint at the best validation epoch under the original activation (GELU or $\lambda$-\acs{gelu}). ``Original'' reports the validation metric at that checkpoint; ``Substituted'' reports the validation metric after replacing the activation by ReLU at the same checkpoint, without further weight updates (mean over $3$ seeds). Overall, \acs{relu}-ization preserves the validation metric substantially better when it is preceded by our annealing procedure: by progressively increasing $\lambda$, the network adapts to harder gating during training, so that the subsequent replacement of $\lambda$-\acs{gelu} by \acs{relu} induces only a minor change in the effective input--output mapping. Across the tabular and vision benchmarks, the annealed replacement consistently reduces the performance drop relative to a direct \acs{gelu}$\rightarrow$\acs{relu} swap, and in some cases it even matches (or slightly improves upon) the pre-replacement score (e.g., Adult). In contrast, directly swapping \acs{gelu} for \acs{relu} without prior hardening typically causes a noticeable degradation, consistent with an abrupt distributional shift in intermediate activations.

Transformer training is markedly more sensitive to activation replacement. For GPT-2 on WikiText-2 (where the validation metric is perplexity, lower is better), replacing \acs{gelu} with \acs{relu} causes a severe deterioration in both the \acs{gelu} and $\lambda$-\acs{gelu} cases, suggesting that additional mechanisms (e.g., post-swap fine-tuning or architecture-specific hardening strategies) are required for language modeling; therefore, we do not report those results here and we leave this analysis for future work.

For Table~\ref{tab:eval}, ``Original'' refers to the validation metric at the selected best-validation checkpoint under the activation used in each row: the \ac{gelu} checkpoint for the \acs{gelu} baseline, and the annealed $\lambda$-\acs{gelu} checkpoint for the $\lambda$-\acs{gelu} row.
\begin{table}[htpb]
\caption{Annealing and replacement evaluation on the validation set.}
\label{tab:eval}
\centering
\begin{adjustbox}{max width=\linewidth}
\begin{tabular}{cccc}
\toprule
\textbf{Dataset} & \textbf{Mode} & \makecell{\textbf{Original}\\\textbf{metric}} & \makecell{\textbf{Substituted}\\\textbf{metric}} \\
\midrule

\multirow{2}{*}{\textbf{Adult}}
& \acs{gelu}         & 0.85  & 0.80 \\
\cmidrule(lr){2-4}
& $\lambda$-\acs{gelu}          & 0.81  & 0.81 \\
\midrule

\multirow{2}{*}{\textbf{FMNIST}}
& \acs{gelu}         & 0.86  & 0.83 \\
\cmidrule(lr){2-4}
& $\lambda$-\acs{gelu}          & 0.86  & 0.85 \\
\midrule

\multirow{2}{*}{\textbf{CIFAR10}}
& \acs{gelu}         & 0.95  & 0.92 \\
\cmidrule(lr){2-4}
& $\lambda$-\acs{gelu}          & 0.95  & 0.95 \\
\midrule

\multirow{2}{*}{\textbf{CIFAR100}}
& \acs{gelu}         & 0.74  & 0.67 \\
\cmidrule(lr){2-4}
& $\lambda$-\acs{gelu}          & 0.74  & 0.72 \\
\midrule

\multirow{2}{*}{\textbf{TinyImageNet}}
& \acs{gelu}         & 0.71  & 0.68 \\
\cmidrule(lr){2-4}
& $\lambda$-\acs{gelu}          & 0.71  & 0.72 \\
\bottomrule
\end{tabular}
\end{adjustbox}
\end{table}

\section{Conclusion and future directions}
\label{sec:conclusions}
We introduced $\lambda$-\acs{gelu}, a hardness-parameterized variant of \acs{gelu} of the form $x\,\Phi(\lambda x)$ with $\lambda\ge 1$, and used it as a lightweight probe to study and control gating hardness in deep networks. To enable stable learning of $\lambda$, we proposed a constrained optimization scheme based on a temperature-controlled softplus reparameterization, and a dedicated learning-rate multiplier for the unconstrained hardness variable. Across a grid search over $(t,c)$, we found that hardness adaptation can be made substantial while keeping the best validation score close to the \acs{gelu} baseline.

Across six model--dataset benchmarks spanning different models (\ac{mlp}, ResNet-18, DeiT, and GPT-2), we observed structured layerwise hardness dynamics and evaluated their robustness to different initializations through Spearman rank correlation. In several vision and tabular settings, the relative ordering of layers by hardness becomes consistent across initializations, suggesting that training induces a repeatable hardness allocation across depth. Finally, we proposed a gradual \acs{relu}-ization procedure: after an empirically selected switch point at $25\%$ of training epochs, we anneal $\lambda$ toward a target $\lambda_{\mathrm{target}}$ chosen from an $\ell_1$ approximation bound between $\Phi(\lambda x)$ and the Heaviside gate. The resulting hardened models can be replaced by \acs{relu} with substantially smaller degradation than a direct \acs{gelu}$\rightarrow$\acs{relu} swap, and in some cases the replacement matches the pre-swap validation metric. Transformer activation replacement remains challenging: while annealing mitigates the shock relative to a direct swap, it does not fully preserve performance in language modeling. We view this as a first step that motivates future work on architecture-aware hardening and replacement criteria in the presence of residual connections and normalization. Our choice of $\lambda_{\mathrm{target}}$ is intentionally distribution-agnostic: it guarantees a uniform gate-approximation tolerance independently of the pre-activation statistics, at the cost of being conservative for normalized architectures. A tighter alternative would set layer-specific targets from empirical pre-activation distributions; we leave this refinement for future work.

This work provides a minimal and practical bridge between smooth-gated training and \acs{relu}-compatible models. Future work includes extending hardness learning to other self-gated activations, exploring finer granularities (per-channel or per-neuron hardness), and developing adaptive schedules for the annealing onset and rate. A particularly important direction is to quantify downstream benefits in the settings that motivate \acs{relu}-ization, such as integer-only quantization, sparsity-based compression, and piecewise-linear analysis and verification pipelines.

Finally, while we motivate \acs{relu}-ization by compatibility with piecewise-linear deployment and analysis pipelines, a systematic evaluation of downstream gains (e.g., integer-only quantization, sparsity-driven pruning, and verification/interpretability toolchains) is an important next step beyond the scope of this submission.
\section*{Acknowledgment}
This research was funded by the projects PID2023-146569NB-C21 and PID2023-146569NB-C22 supported by MICIU/AEI/10.13039/501100011033 and ERDF/UE. Cristian Pérez-Corral received support from the \textit{Conselleria de Educación, Cultura, Universidades y Empleo} (reference CIACIF/2024/412) through the European Social Fund Plus 2021–2027 (FSE+) program of the \textit{Comunitat Valenciana}. Alberto Fernández-Hernández was supported by the predoctoral grant PREP2023-001826 supported by MICIU/AEI/10.13039/501100011033 and ESF+. Jose I. Mestre was supported by the predoctoral grant ACIF/2021/281 of the \emph{Generalitat Valenciana}. Manuel F. Dolz was supported by the Plan Gen--T grant CIDEXG/2022/13 of the \emph{Generalitat Valenciana}

\balance
\bibliographystyle{IEEEtran}
\bibliography{bibliography}

@inproceedings{pmlr-v15-glorot11a,
  title     = {Deep Sparse Rectifier Neural Networks},
  author    = {Glorot, Xavier and Bordes, Antoine and Bengio, Yoshua},
  booktitle = {Proceedings of the Fourteenth International Conference on Artificial Intelligence and Statistics},
  year      = {2011},
  series    = {Proceedings of Machine Learning Research},
  publisher = {PMLR},}

@inproceedings{DBLP:conf/nips/KrizhevskySH12,
  author    = {Alex Krizhevsky and Ilya Sutskever and Geoffrey E. Hinton},
  editor    = {Peter L. Bartlett and Fernando C. N. Pereira and Christopher J. C. Burges and L{\'{e}}on Bottou and Kilian Q. Weinberger},
  title     = {ImageNet Classification with Deep Convolutional Neural Networks},
  booktitle = {Advances in Neural Information Processing Systems 25: 26th Annual Conference on Neural Information Processing Systems 2012},
  year      = {2012}
}

@article{elfwing2017silu,
  author  = {Elfwing, Stefan and Uchibe, Eiji and Doya, Kenji},
  title   = {Sigmoid-Weighted Linear Units for Neural Network Function Approximation in Reinforcement Learning},
  year    = {2017},
  journal = {arXiv preprint}
}

@inproceedings{chu2018,
author = {Chu, Lingyang and Hu, Xia and Hu, Juhua and Wang, Lanjun and Pei, Jian},
title = {Exact and Consistent Interpretation for Piecewise Linear Neural Networks: A Closed Form Solution},
year = {2018},
booktitle = {Proceedings of the 24th ACM SIGKDD International Conference on Knowledge Discovery \& Data Mining},
series = {KDD '18}
}

@article{hendrycks2016gaussian,
  title={Gaussian error linear units (gelus)},
  author={Hendrycks, Dan and Gimpel, Kevin},
  journal={arXiv preprint arXiv:1606.08415},
  year={2016}
}

@misc{ramachandran2017searchingactivationfunctions,
      title={Searching for Activation Functions}, 
      author={Prajit Ramachandran and Barret Zoph and Quoc V. Le},
      year={2017},
}

@misc{misra2020mishselfregularizednonmonotonic,
      title={Mish: A Self Regularized Non-Monotonic Activation Function}, 
      author={Diganta Misra},
      year={2020}
}

@inproceedings{nair2010relu,
author = {Nair, Vinod and Hinton, Geoffrey E.},
title = {Rectified linear units improve restricted boltzmann machines},
year = {2010},
publisher = {Omnipress},
address = {Madison, WI, USA},
booktitle = {Proceedings of the 27th International Conference on International Conference on Machine Learning},
series = {ICML'10}
}

@article{hu2016,
  author       = {Hengyuan Hu and
                  Rui Peng and
                  Yu{-}Wing Tai and
                  Chi{-}Keung Tang},
  title        = {Network Trimming: {A} Data-Driven Neuron Pruning Approach towards
                  Efficient Deep Architectures},
  journal      = {CoRR},
  volume       = {abs/1607.03250},
  year         = {2016},
  }

@article{kim2021bert,
  title={I-BERT: Integer-only BERT Quantization},
  author={Kim, Sehoon and Gholami, Amir and Yao, Zhewei and Mahoney, Michael W and Keutzer, Kurt},
  journal={International Conference on Machine Learning},
  year={2021}
}

@InProceedings{jacob2018,
author = {Jacob, Benoit and Kligys, Skirmantas and Chen, Bo and Zhu, Menglong and Tang, Matthew and Howard, Andrew and Adam, Hartwig and Kalenichenko, Dmitry},
title = {Quantization and Training of Neural Networks for Efficient Integer-Arithmetic-Only Inference},
booktitle = {Proceedings of the IEEE Conference on Computer Vision and Pattern Recognition (CVPR)},
month = {June},
year = {2018}
}

@article{katz2017,
  author       = {Guy Katz and
                  Clark W. Barrett and
                  David L. Dill and
                  Kyle Julian and
                  Mykel J. Kochenderfer},
  title        = {Reluplex: An Efficient {SMT} Solver for Verifying Deep Neural Networks},
  journal      = {CoRR},
  volume       = {abs/1702.01135},
  year         = {2017}
}

@inproceedings{katz2019,
author = {Katz, Guy and Huang, Derek and Ibeling, Duligur and Julian, Kyle and Lazarus, Christopher and Lim, Rachel and Shah, Parth and Thakoor, Shantanu and Wu, Haoze and Zeljić, Aleksandar and Dill, David and Kochenderfer, Mykel and Barrett, Clark},
year = {2019},
title = {The Marabou Framework for Verification and Analysis of Deep Neural Networks},
booktitle="Computer Aided Verification"
}

@inproceedings{he2015delving,
  title={Delving deep into rectifiers: Surpassing human-level performance on imagenet classification},
  author={He, Kaiming and Zhang, Xiangyu and Ren, Shaoqing and Sun, Jian},
  booktitle={Proceedings of the IEEE international conference on computer vision},
  pages={1026--1034},
  year={2015}
}

@article{cifar10,
author = {Krizhevsky, Alex},
year = {2012},
month = {05},
pages = {},
title = {Learning Multiple Layers of Features from Tiny Images},
journal = {University of Toronto}
}

@INPROCEEDINGS{resnet,
  author={He, Kaiming and Zhang, Xiangyu and Ren, Shaoqing and Sun, Jian},
  booktitle={2016 IEEE Conference on Computer Vision and Pattern Recognition (CVPR)}, 
  title={Deep Residual Learning for Image Recognition}, 
  year={2016},
  }

@misc{adult,
  author       = {Becker, Barry and Kohavi, Ronny},
  title        = {{Adult}},
  year         = {1996},
  howpublished = {UCI Machine Learning Repository},
  note         = {{DOI}: https://doi.org/10.24432/C5XW20}
}

@misc{fmnist,
      title={Fashion-MNIST: a Novel Image Dataset for Benchmarking Machine Learning Algorithms}, 
      author={Han Xiao and Kashif Rasul and Roland Vollgraf},
      year={2017}
}

@misc{deit,
      title={Training data-efficient image transformers \& distillation through attention}, 
      author={Hugo Touvron and Matthieu Cord and Matthijs Douze and Francisco Massa and Alexandre Sablayrolles and Hervé Jégou},
      year={2021}
}

@article{gpt2,
  title={Language Models are Unsupervised Multitask Learners},
  author={Radford, Alec and Wu, Jeff and Child, Rewon and Luan, David and Amodei, Dario and Sutskever, Ilya},
  year={2019},
  journal = {OpenAI}
}

@inproceedings{
wikitext,
title={Pointer Sentinel Mixture Models},
author={Stephen Merity and Caiming Xiong and James Bradbury and Richard Socher},
booktitle={International Conference on Learning Representations},
year={2017}
}

@inproceedings{tiny,
  title={Tiny ImageNet Visual Recognition Challenge},
  author={Ya Le and Xuan S. Yang},
  year={2015},
  booktitle = {}
}
\end{document}